%% file: main.tex
\documentclass[10pt,twocolumn,letterpaper]{article}

\usepackage[pagenumbers]{cvpr} % To force page numbers, e.g. for an arXiv version

\input{preamble}
\definecolor{cvprblue}{rgb}{0.21,0.49,0.74}
\usepackage[pagebackref,breaklinks,colorlinks,allcolors=cvprblue]{hyperref}
\usepackage{multirow} 
\def\confName{CVPR}
\def\confYear{2026}

\title{\vspace{-1cm}RoboVIP: Multi-View Video Generation with Visual Identity Prompting \\Augments Robot Manipulation}

\author{
Boyang Wang\textsuperscript{1}\thanks{ Indicates co–first author.},
Haoran Zhang\textsuperscript{4}\footnotemark[1],
Shujie Zhang\textsuperscript{1,2}\footnotemark[1],
Jinkun Hao\textsuperscript{1,3},
Mingda Jia\textsuperscript{1},\\
Qi Lv\textsuperscript{1},
Yucheng Mao\textsuperscript{1},
Zhaoyang Lyu\textsuperscript{1},
Jia Zeng\textsuperscript{1},
Xudong Xu\textsuperscript{1}\thanks{ Corresponding author.},
Jiangmiao Pang\textsuperscript{1}\\[4pt]
\textsuperscript{1}Shanghai AI Laboratory \quad
\textsuperscript{2}Tsinghua University \quad \\
\textsuperscript{3}Shanghai Jiao Tong University\quad 
\textsuperscript{4}University of Michigan \\
\textcolor{blue}{Homepage: \url{https://robovip.github.io/RoboVIP/}}
}

\begin{document}
\maketitle
\input{sec/0_abstract}    
\input{sec/1_intro}

\input{sec/2_related_work}

\input{sec/3_method}

\input{sec/4_experiment}

\input{sec/5_conclusion}

{
    \small
    \bibliographystyle{ieeenat_fullname}
    \bibliography{main}
}

\clearpage

\input{sec/X_suppl}

\end{document}

%% file: sec/0_abstract.tex
\begin{abstract}

The diversity, quantity, and quality of manipulation data are critical for training effective robot policies. However, due to hardware and physical setup constraints, collecting large-scale real-world manipulation data remains difficult to scale across diverse environments. Recent work uses text-prompt conditioned image diffusion models to augment manipulation data by altering the backgrounds and tabletop objects in the visual observations. However, these approaches often overlook the practical need for multi-view and temporally coherent observations required by state-of-the-art policy models. Further, text prompts alone cannot reliably specify the scene setup. To provide the diffusion model with explicit visual guidance, we introduce visual identity prompting, which supplies exemplar images as conditioning inputs to guide the generation of the desired scene setup. To this end, we also build a scalable pipeline to curate a visual identity pool from large robotics datasets.  Using our augmented manipulation data to train downstream vision-language-action and visuomotor policy models yields consistent performance gains in both simulation and real-robot settings.

\end{abstract}

%% file: sec/1_intro.tex
\section{Introduction}
\label{sec:intro}

High-quality and diverse visual data remains fundamental to progress in robotic manipulation and policy learning~\cite{team2024octo, black2024pi_0, kim2024openvla}.
However, collecting such data in the real world is notoriously challenging: each episode requires precise mechanical setups, calibrated camera rigs, and reliable synchronization across sensing devices.
These constraints make it difficult to scale manipulation datasets in both quantity and environmental diversity.
As a complementary solution, recent work has turned to generative models~\cite{ho2020denoising, song2020score} to synthesize additional data, offering a promising alternative to labor-intensive data collection~\cite{yu2023scaling, chen2025semantically, yuan2025roboengine}.

A growing line of work~\cite{yuan2025roboengine, yu2023scaling, chen2025semantically} augments visual observations in manipulation data while keeping the underlying action trajectory fixed.
% These approaches typically operate in a single-frame, single-view setting: 
They segment the robot arm and the interacted objects, then apply text prompt-guided image generative models~\cite{rombach2022high} to \textbf{\emph{inpaint}} masked regions, diversifying backgrounds and table-top contents. 
However, these approaches typically operate in a single-frame, single-view setting, which diverges from the needs of modern policy models~\cite{chi2025diffusion, kim2024openvla, black2024pi_0, team2024octo}.
First, many manipulation tasks inherently require reasoning over longer temporal histories, rather than relying solely on a single observation.
Consider a policy model executing a \textit{push-button} task where the pre-interaction and post-interaction states of the button appear visually identical. 
With only one historical frame, the policy model cannot tell whether it has already pushed the button or is about to, often producing indecisive behaviors, even action loops.
Second, multi-view observations are increasingly adopted in visuomotor policy models~\cite{chi2025diffusion, zhao2023learning} and VLA systems~\cite{black2024pi_0, kim2024openvla, team2024octo}, and are usually provided in recent robotics datasets~\cite{khazatsky2024droid, ebert2021bridge}, as they provide richer spatial cues and better cross-view generalization.
As a result, a practical augmentation framework should operate at the \textbf{\emph{video level}} and support \textbf{\emph{multi-view}} generation.

\begin{figure*}[t]
\centering
% \maskplaceholder{0.95\textwidth}{55mm}{}
\vspace{-0.3cm}
\includegraphics[width=0.95\textwidth]{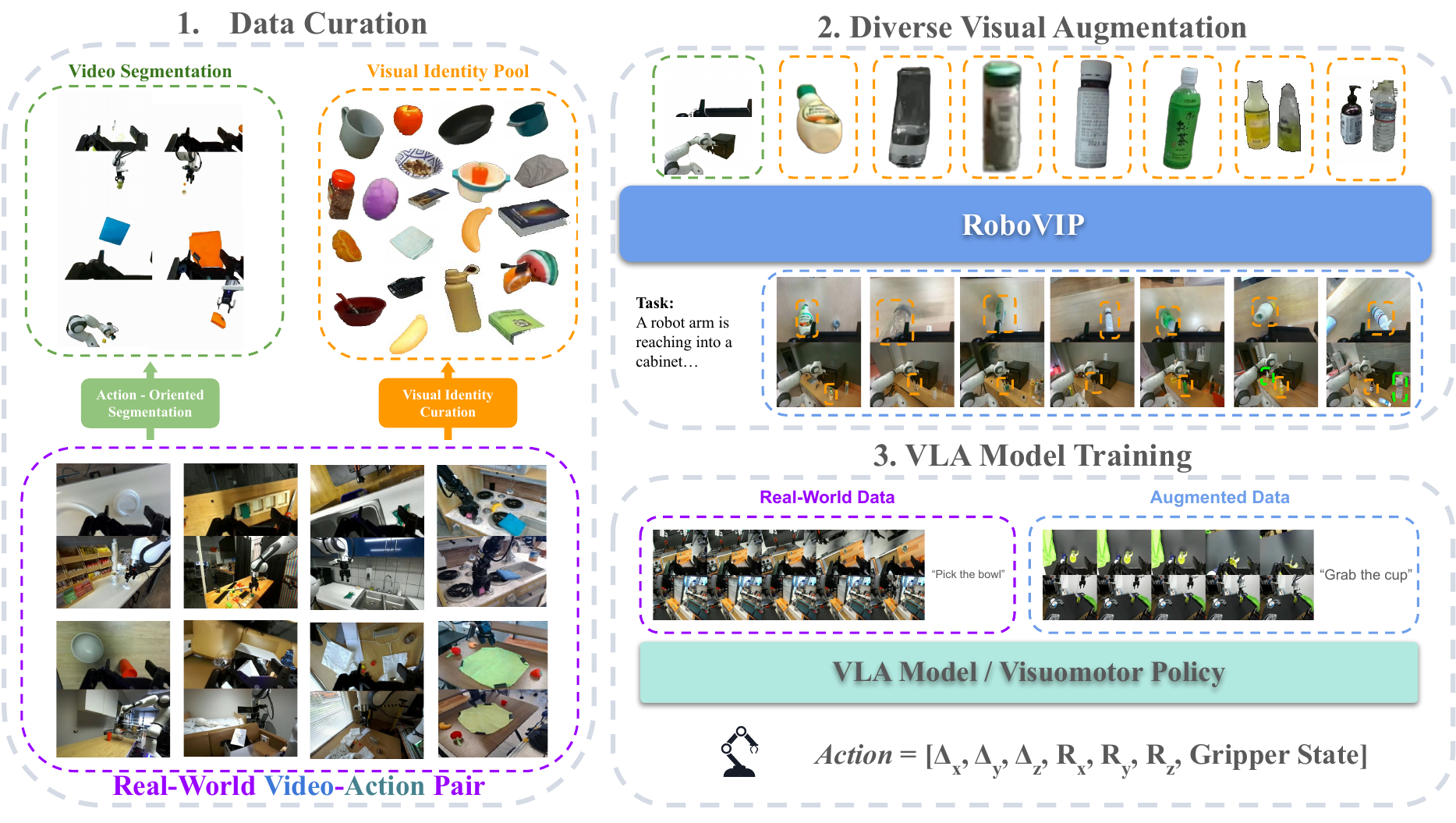}
\vspace{-0.3cm}
\caption{
    \textbf{Overview of our RoboVIP Workflow.} 
    (1) We extract observation videos from robotics manipulation data with corresponding action data to segment the robot arm and interacted objects for inpainting-based augmentation.
    (2) A large-scale pool of visual identity prompts is curated from robotics datasets and used as conditioning inputs for our multi-view video diffusion model to conduct diverse visual augmentation.
    (3) The augmented videos, paired with action information from original robotics manipulation data, are utilized for downstream VLA and visuomotor policy training.
}
\vspace{-0.25cm}
\label{fig:main_pipeline}
\end{figure*}

In this work, we present a multi-view video generation augmentation framework—featuring dynamically moving wrist-camera views—that enables diversification of backgrounds and tabletop scenes in a fully \textbf{\emph{plug-and-play}} manner using only raw videos as input.
This framework necessitates an automatic segmentation pipeline to segment out both the robot and the objects being interacted with.
In practice, the interacted object may be invisible in early wrist-camera frames, coupled with rapid camera motion, narrow field of view, long trajectories, and limited robotics-specific training; thus, directly applying off-the-shelf models like vision-language models~\cite{bai2025qwen2, azzolini2025cosmos} (VLM) often fail to localize the target object reliably.
To address these issues, we propose an automated segmentation pipeline that leverages action information to mitigate segmentation failures from the off-the-shelf model, especially on the wrist-camera.
Specifically, we use the 1D gripper state to identify the time window that the robot arm actually interacts with the target object, which narrows the search space.

Furthermore, we observe that solely relying on text-prompt–guided generation, as done in prior works~\cite{yu2023scaling, yuan2025roboengine, chen2025semantically, alhaija2025cosmos, ali2025world}, imposes limitations.
The text prompts provided by existing datasets~\cite{khazatsky2024droid, ebert2021bridge} are typically overly simplistic and lack detailed table-top descriptions. 
Even if we apply SoTA VLM~\cite{bai2025qwen2} to caption, the generated descriptions frequently suffer from hallucinations and misalignment.
More importantly, text prompts can not capture low-level details. 
To mitigate these issues, we introduce \textbf{\emph{visual identity prompting}}, conditioning the video diffusion model on one or more exemplar images to synthesize both semantically and low-level consistent content in the inpainted regions.
This conditioning will force the model to enrich the table-top and background contents.
Further, we modify our video diffusion model to incorporate visual identity prompting conditions in the multi-view paradigm and propose an efficient scheme to embrace multiple identities at the same time.
To preserve the plug-and-play nature of our framework, visual identities are \emph{not manually} provided by humans, as in general video generation methods~\cite{liu2025phantom, fei2025skyreels}.
Instead, we propose an agentic curation and filtering pipeline that automatically constructs a million-scale visual identity pool from large-scale robotics datasets.
The full pipeline is shown in Fig.~\ref{fig:main_pipeline}.

In summary, we present \textbf{\emph{RoboVIP}}, a multi-view inpainting-based video diffusion model with visual identity prompting as conditions to augment the visual observations of the robotics manipulation data. 
Our approach integrates an action-oriented pipeline for multi-view robot and object segmentation, a scalable curation pipeline to construct a large-scale visual identity pool, and a video diffusion model capable of generating temporally consistent multi-view sequences with visual identity prompting as conditions.
To assess the effectiveness at scale, we augment 12K BridgeV2~\cite{walke2023bridgedata} trajectories for training mainstream VLA models, including $\pi_0$~\cite{black2024pi_0} and Octo~\cite{team2024octo}. 
We further evaluate RoboVIP on a real-world robot dataset (100 trajectories) for training a visuomotor policy, Diffusion Policy~\cite{chi2025diffusion}. 
Across both simulation and real-world robot evaluations, RoboVIP delivers consistent gains in success rate, demonstrating its practicality for large-scale VLA training as well as low-data policy learning.

%% file: sec/2_related_work.tex
\section{Related Works}
\label{sec:formatting}

\subsection{Conditioned Video Generation}
Video generation synthesizes realistic, temporally coherent sequences conditioned on text, images, or videos~\cite{yang2024cogvideox, wan2025wan, ku2024anyv2v, alhaija2025cosmos}.
Video-to-video models transform an input clip into another consistent sequence, enabling style transfer~\cite{ku2024anyv2v}, enhancement~\cite{zhou2024upscale}, and content editing~\cite{ju2025editverse}.
Beyond pixel-aligned cues, identity references~\cite{liu2025phantom, fei2025skyreels, wang2025frame} have emerged as a way to inject explicit visual attributes into generation.
Although video generation is increasingly used for robot planning~\cite{du2023learning, li2025unified, hu2024video} and controllable video generation is gaining attention~\cite{wang2025language, zhu2025irasim}, its use for visual augmentation remains underexplored: most existing approaches rely on image-based diffusion~\cite{yu2023scaling, yuan2025roboengine} or support only single-view conditioning~\cite{alhaija2025cosmos}, leaving multi-view and richer conditioning largely unexamined.

% \vspace{-0.5\baselineskip}
\subsection{Visual Augmentation on Robotics}
Traditional augmentations like cropping, rotation, and resizing offer limited benefit for robot policy training and do not resolve data scarcity. 
As a result, practitioners resort to applying learning-based methods to increase visual diversity.
GreenAug~\cite{teoh2024green} augments the dataset by manually setting the real-world robot manipulation environment with the green screen and generating the background by post-effects. 
ReBot~\cite{fang2025rebot} and RoboSplat~\cite{yang2025novel} conduct a real-to-sim-to-real paradigm, which converts the real-world robot information to a simulation environment and then manually adds objects, changes views, or poses of the objects to create a hand-crafted dataset.
However, these methods require significant manual effort and do not generalize to new tasks or environments in a plug-and-play manner.
To achieve plug-and-play, Cosmos-Transfer~\cite{alhaija2025cosmos, ali2025world} and RoboTransfer~\cite{liu2025robotransfer} use pixel-aligned conditions, like edges, depth, and segmentation, to drive the diffusion model on appearance-level editing. 
Although effective, this condition design limits the ability to introduce new semantic content for the diffusion model.
Instead, Rosie~\cite{yu2023scaling} and RoboEngine~\cite{yuan2025roboengine} apply masking to the background and table-top contents and then apply a generative model to inpaint masked areas. This unleashes the power of generative models beyond appearance-level editing. 
% Alongside these works, we introduce an accurate multi-view masking pipeline and enable finer controllability over the masked regions by visual identity prompting.

\subsection{Manipulation Models}

Research on robot manipulation has progressed from early visuomotor policies to unified VLA architectures. Classic visuomotor systems \cite{chi2025diffusion, levine2016end} map images to actions with supervised or reinforcement learning; however, task-specific demonstrations limit generalization. Large vision-language models \cite{radford2021learning, alayrac2022flamingo} broadened representational capacity, which in turn motivated VLAs that jointly encode vision, language, and action. Two design axes are now prominent: first, temporal conditioning, where models range from short history windows (RT-1 \cite{rt12022rt1}, Octo\cite{team2024octo}, OpenVLA\cite{team2024octo}) to sequence encoders; second, viewpoint, where many systems use a single egocentric stream, whereas newer ones incorporate multi-view inputs to improve 3D reasoning (e.g., $\pi_0$\cite{black2024pi_0}). As a result, training increasingly requires temporally coherent and cross-view-aligned data, yet such data remain scarce because real-world collection is slow, costly, and rarely long-horizon or synchronized \cite{brohan2023rtx}. This gap motivates data augmentation that generates additional supervision while preserving frame-to-frame and view-to-view consistency.

%% file: sec/3_method.tex
\section{Method}

\subsection{Problem Formulation}

As shown in Fig.~\ref{fig:main_pipeline}, starting from a robotic manipulation episode with multi-view video sequences and its corresponding action information on the end-effector state, we segment the robot arm and the objects being interacted with to preserve the fundamental 6-DoF Cartesian end-effector delta pose and gripper-state information. 
For the remaining masked regions---including background, foreground, and tabletop objects---we adopt a video diffusion model that is conditioned on the text prompt $\mathbf{y}$, the masked multi-view videos $\mathbf{M} = \{\textit{M}_0, ..., \textit{M}_N\}$, and the proposed visual identity prompting $\mathbf{f} = \{\mathbf{f_1}, ..., \mathbf{f_k}\}$. 
The model is trained to learn the conditional joint distribution
$p_{\theta}(\textit{I}_0, ..., \textit{I}_N \mid \textit{M}_0, ..., \textit{M}_N, y, f_1, ..., f_k),$
where $\mathbf{I} = \{\textit{I}_0, ..., \textit{I}_N\}$ denotes the generated latent video frames that adhere to all conditioning signals.
The inpainted multi-view videos serve as augmented observations for policy model training, while the original action sequences are directly reused.

\subsection{Action-guided Segmentation of Robots and Interacted Objects}

\begin{figure}[t]
\centering
% \maskplaceholder{0.95\textwidth}{55mm}{}
% \vspace{-0.1cm}
\includegraphics[width=0.47\textwidth]{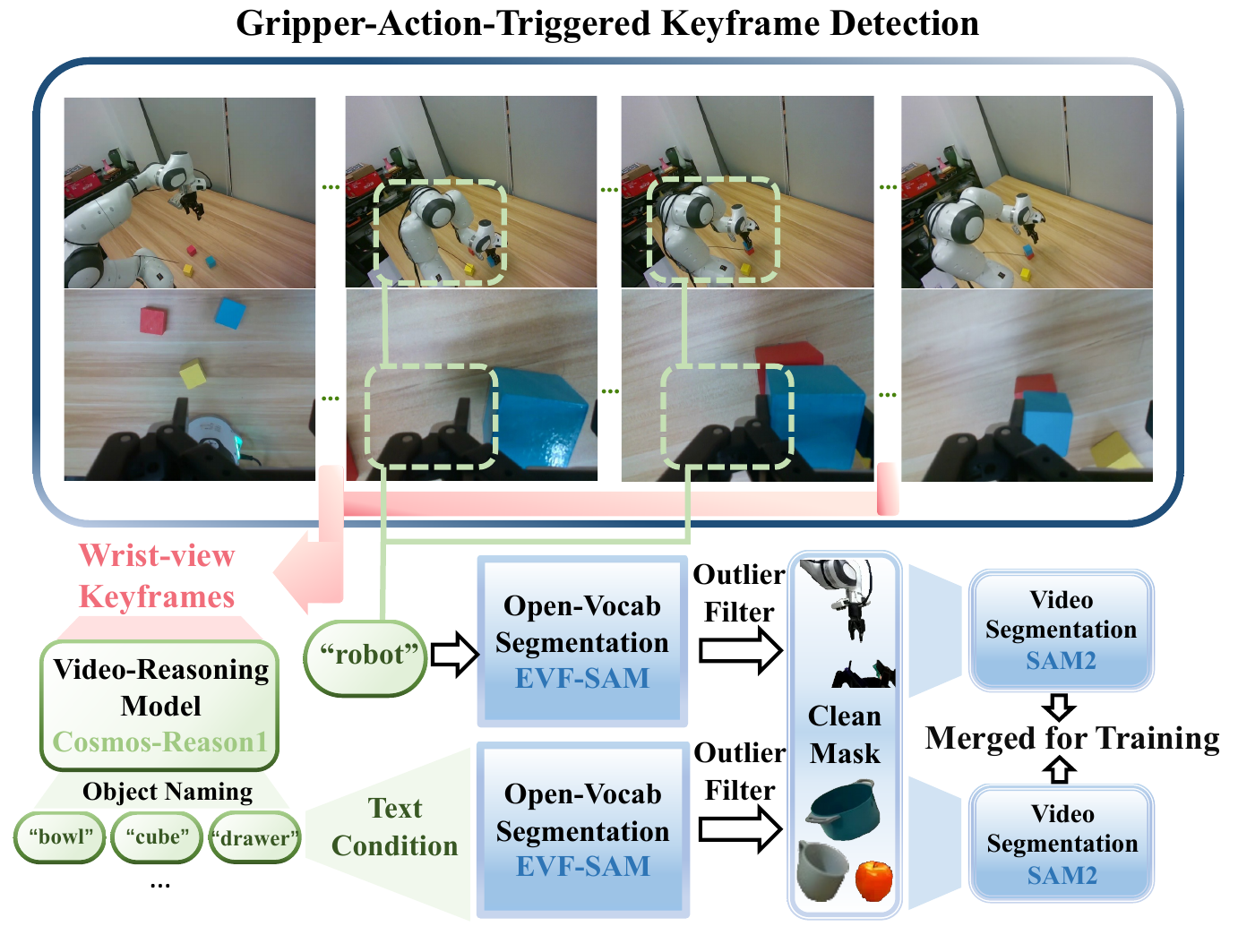}
\vspace{-0.3cm}
\caption{
    \textbf{Segmentation Pipeline.}
    Our segmentation pipeline comprises two parallel streams: one for robot-arm segmentation and one for interacted-object segmentation. We first use the gripper-action signal to identify accurate keyframe ranges, which is helpful to locate the interacted objects that are not visible in the first or last frame. We then leverage off-the-shelf models such as Cosmos-Reason1~\cite{azzolini2025cosmos} and SAM2~\cite{ravi2024sam}, together with several heuristic refinements, to obtain accurate masks in a fully plug-and-play manner.
    }
\label{fig:seg_pipeline}
\vspace{-0.4cm}
\end{figure}

The action information generally includes the 6-DoF pose of the end-effector $\Delta x$, $\Delta y$, $\Delta z$, $\Delta roll$, $\Delta pitch$, $\Delta yaw$, and a 1-D gripper state. The gripper state indicates when the robot arm closes or opens, which provides decisive cues for most robot manipulation tasks. 
In a long video sequence, an effective grasp by the robot arm occurs only within a very short time window, and the moment when the gripper state changes can be used to localize the interacted object from the wrist-view perspective. 
Specifically, as shown in Fig.~\ref{fig:seg_pipeline}, we first identify the frame range corresponding to gripper-closure intervals in the wrist view, which marks the preparation and execution of interaction. 
The extracted video clip is then fed into a video-reasoning VLM~\cite{azzolini2025cosmos} to infer the object's semantic label, enabling accurate object naming directly from the wrist view.

When processing other third-person camera views, the object name obtained from the wrist view is directly reused. The identified object name is fed into an open-vocabulary segmentation model~\cite{yuan2025roboengine, zhang2024evf} to obtain a reliable mask for the corresponding frame. We separately extract the masks for the robot and the interacted object, followed by median blurring to filter out outlier pixels. 
At this stage, we locate accurate temporal range and mask locations. To further refine temporal consistency, we perform k-means sampling on the masks, and the sampled points will be taken as prompting for the video segmentation model~\cite{ravi2024sam} to track the complete video segmentation for the robot and interacted objects. The robot and the object mask are processed independently and then merged into one at the end. 
The resulting video segmentation provides high-quality mask conditions that can be directly used in the training process.

\subsection{Multi-View Inpainting Video Diffusion Model}

\begin{figure}[t]
\centering
\vspace{-0.2cm}
\includegraphics[width=0.45\textwidth]{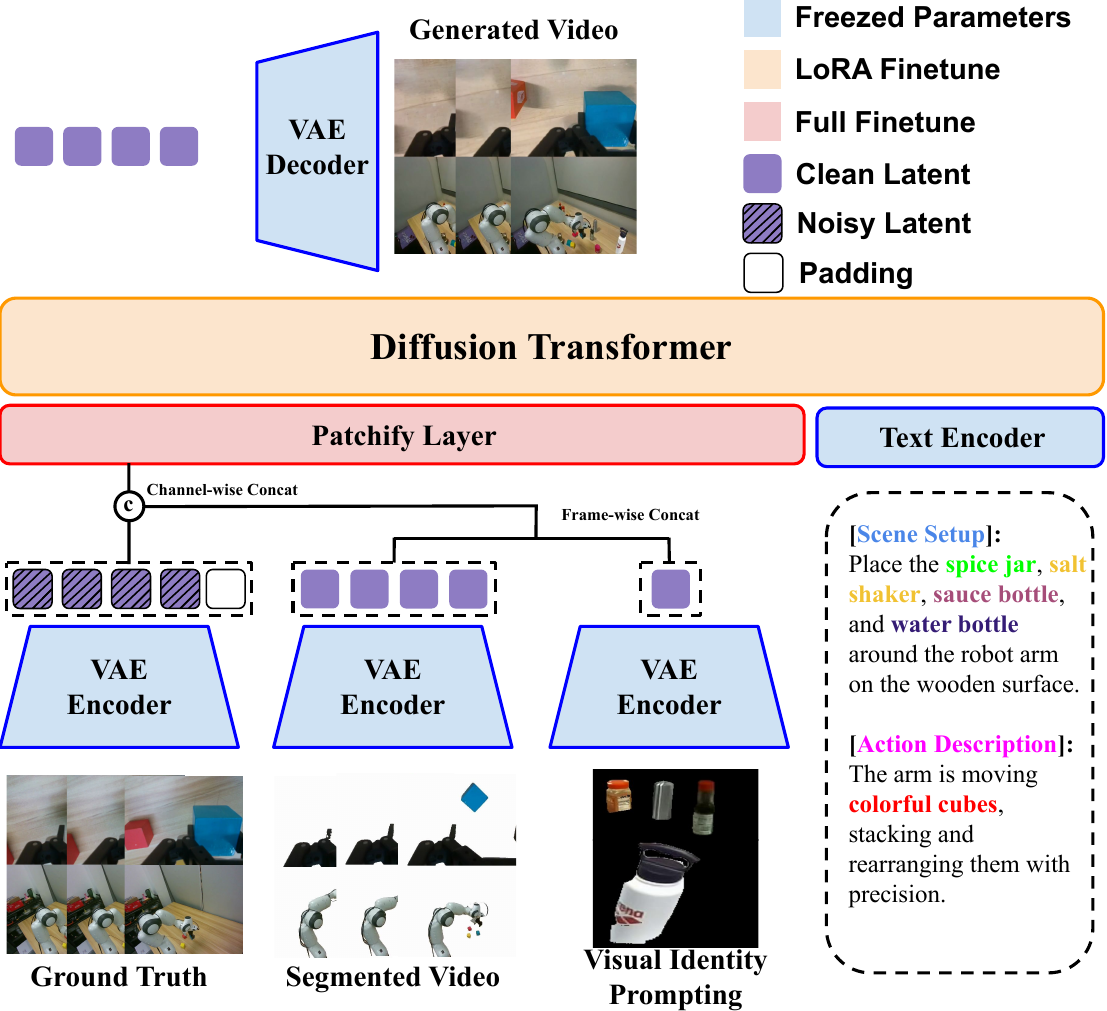}
\vspace{-0.2cm}
\caption{
    \textbf{Video Diffusion Model Architecture.} Our video diffusion model is conditioned on the segmented multi-view video sequence, structured text prompt, and visual identity prompting to achieve consistent visual augmentation.
}
\label{fig:vdm_figure}
\vspace{-0.5cm}
\end{figure}

We aim to transfer the generation quality and condition-alignment capability of state-of-the-art video diffusion models to robotic tasks. To this end, our base model is the Wan2.1~\cite{wan2025wan} image-to-video variant with 14 billion parameters. 
However, directly fine-tuning such a large model is computationally infeasible. More critically, it leads to severe overfitting collapse, causing the model to rapidly forget its original visual generation stability. 
To address this, we adopt a Low-Rank Adaptation~\cite{hu2022lora} (LoRA) strategy to enable feasible and memory-efficient fine-tuning.

Recent video diffusion models are predominantly built upon the Diffusion Transformer~\cite{peebles2023scalable} architecture, where attention blocks serve as the primary computational units. 
LoRA injects trainable low-rank adapters into the linear projections, 
typically applied to the query and value matrices within attention layers.
Apart from the attention blocks, the patchification layer—implemented as a convolutional layer that transforms latent images into patches—is fully trainable but not part of LoRA fine-tuning.
Since our training objective shifts from single-image conditioning in the base model to the masked video sequences as conditions, we enable the patchification encoder for training as well. 
Empirically, we find that fine-tuning this additional layer beyond the LoRA setup tends to yield slight improvements in performance.

For multi-view inputs, inspired by~\cite{jang2025dreamgen}, we adopt a structured vertical stitching strategy, which concatenates masked frames from different views at the same timestamp.
The ground-truth sequences are processed in the same manner, ensuring that the learning objective remains view-aligned and encourages the video diffusion model to capture cross-view spatial consistency and correspondence.
% an elegant yet stable approach to handle multi-view inputs. 
Accordingly, we modify the base model’s input structure by replacing the single-image padding with channel-wise concatenation of the full video sequence, which achieves a minimally invasive yet effective formulation of a video-conditioned objective.
The overall model structure can be found in Fig.~\ref{fig:vdm_figure}. 
% The visual identity prompting part is in Sec.~\ref{sec:method_visual_identity_prompting}.

\begin{figure}[t]
\centering
% \maskplaceholder{0.45\textwidth}{55mm}{Placeholder}
\vspace{-0.2cm}
\includegraphics[width=0.45\textwidth]{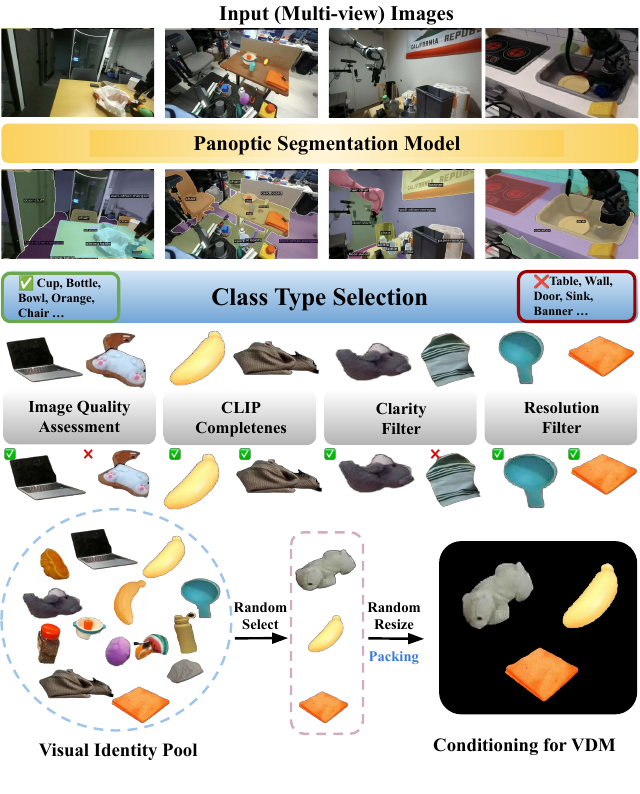}
\vspace{-0.6cm}
\caption{
    \textbf{Visual Identity Curation and Processing Pipeline.} 
    Our visual identity is curated by panoptic segmentation from the large-scale robotics dataset~\cite{ebert2021bridge, walke2023bridgedata, khazatsky2024droid}, followed by several scoring criteria filters. In augmentation, we randomly select some from the pool and pack them into one image frame to serve as conditioning for our video diffusion model.
}
\vspace{-0.5cm}
\label{fig:id_curation_pipeline}
\end{figure}

\subsection{Visual Identity Prompting}
\label{sec:method_visual_identity_prompting}

For robotic downstream tasks, we aim for a pipeline that can autonomously select appropriate and necessary visual identities without any human intervention. To achieve this, we design an agentic inference pipeline, as shown in Fig.~\ref{fig:id_curation_pipeline}, that automatically constructs a massive, rich, and diverse visual identity pool. We find that adopting a panoptic segmentation~\cite{kirillov2019panoptic} approach is the most straightforward way to achieve this goal. Panoptic segmentation simultaneously provides mask localization and corresponding label classification. 
Based on the classification label, we select common objects that are needed and do not consider background-related large objects, like the table and the wall.
Using these labels, we can naturally classify both tabletop objects and background elements, ultimately forming a comprehensive visual identity pool.
To this end, we constitute a million-scale visual identity pool.

We observe that objects obtained by straightforward segmentation are often of suboptimal quality. 
Moreover, some of these segmented objects are partially occluded and thus cannot serve as semantically complete visual identity references.
To address this, we crop the corresponding visual identity image predicted by the panoptic segmentation model~\cite{jain2023oneformer} and then apply several filtering criteria, including image quality assessment~\cite{wang2023exploring}, sharpness clarity assessment, CLIP-based text–image scoring~\cite{radford2021learning}, and resolution size filtering. 
The CLIP text embedding is derived from the panoptic segmentation class label, serving as an effective proxy to assess the semantic completeness of each object.

Unlike previous approaches that inject only a single identity reference per frame, we adopt a packing scheme to efficiently accommodate multiple visual identity references within a single frame, thereby reducing computational overhead.
To prevent overfitting to fixed scale ratios, each identity image is randomly resized before encoding.
During training under multi-view supervision, all visual identity references are sampled from a single view to avoid view-ambiguity in identity prompting.

To incorporate visual identity prompting into the video diffusion model, we adopt a frame-wise concatenation strategy, following the design of~\cite{wang2025frame, zhong2025concat}.
As illustrated in Fig.~\ref{fig:vdm_figure}, before entering the video diffusion transformer, the packed identity images are first encoded by a shared causal VAE encoder~\cite{wan2025wan} and concatenated with the latent video segmentation inputs along the frame dimension. The noisy frame latent is zero-padded for temporal alignment and then channel-wise concatenated with the conditional inputs.
After the diffusion transformer processes all layers, the identity tokens are dropped and excluded from loss computation, ensuring they serve purely as contextual guidance rather than optimization targets.
During inference, newly encoded identity images are injected at each diffusion timestep to continuously guide generation.

%% file: sec/4_experiment.tex
\section{Experiment}

\begin{figure*}[t]
\centering
\vspace{-0.4cm}
\includegraphics[width=0.98\textwidth]{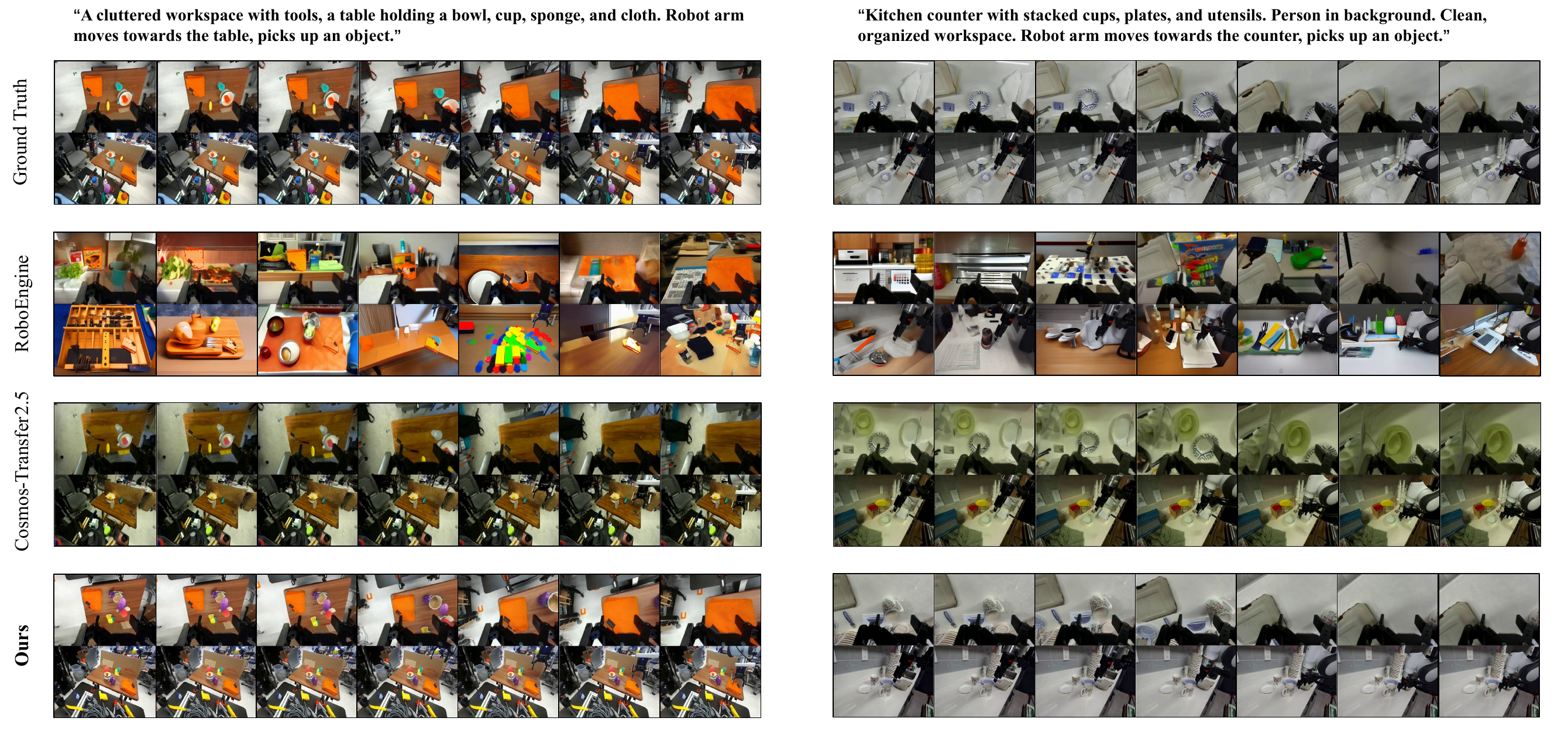}
\vspace{-0.45cm}
\caption{
\textbf{Qualitative comparisons of different models on Droid~\cite{khazatsky2024droid}}.
Our method produces temporally consistent and visually diverse results, outperforming RoboEngine~\cite{yuan2025roboengine}, which is a single-image-based method, and Cosmos-Transfer2.5~\cite{ali2025world}, which struggles to generalize beyond appearance-level edge conditioning. 
\textbf{Zoom in} for the best view.
}
\vspace{-0.3cm}
\label{fig:vdm_comparison}
\end{figure*}

\subsection{Video Diffusion Model Implementation Details}

\textbf{Data Curation.}
For all videos, we first discard sequences that are too short (fewer than 25 frames). For overly long sequences (more than 550 frames), we perform temporal cropping to mitigate segmentation failures induced by excessively extended inputs.
Since the captions provided by the original dataset are often noisy, we re-caption all videos using Qwen2.5-VL 32B~\cite{bai2025qwen2}, employing a multi-view vertical stitching strategy to ensure consistent and accurate textual descriptions across different viewpoints. 
The text prompt is composed of the scene setup and the action description.
Then, for the robot and object segmentation described in the method section, we adopt OneFormer~\cite{jain2023oneformer} for the panoptic segmentation. 
The open-vocab segmentation model is the EVF-SAM~\cite{zhang2024evf} model from RoboEngine~\cite{yuan2025roboengine}. The video segmentation is done by SAM2~\cite{ravi2024sam}.

\noindent\textbf{Training Details.} 
We train our video diffusion models on Bridge V1~\cite{ebert2021bridge} and V2~\cite{walke2023bridgedata} for the following downstream VLA tasks, which provide one to three third-person views.
Further, we train Droid~\cite{khazatsky2024droid} for the visual quality comparisons and real-robot augmentation, which includes a wrist-mounted camera and two third-person views.
To support variable-length sequences, we adopt a batch size of 1 per GPU and use gradient accumulation to achieve an effective batch size of 4 per GPU, which costs around 70GB per GPU in training. This strategy enables dynamic frame sampling without incurring unnecessary computation from padding or attention mask overhead.
Since current VLA models cannot condition on long observation histories, we train on at most 49 frames.
We train for 15K iterations on 8 GPUs whose per-GPU memory is 144GB, resulting in a total cumulative batch size of 32. 
Each view is trained at 256×256 resolution for Bridge and 320×416 for Droid.
When an instance contains only a single view, the conditioning input for the missing view is zero-padded with black pixels to distinguish 255-value white pixels of the segmentation masks. 
% This is because we want to avoid ambiguity between zero-padding and segmentation masks.

In practice, we observe that generative quality correlates with the model’s pretrained resolution; therefore, the cumulative stitched width and height for multi-view must be lower than the pretrained setting. Guided by this finding, we employ the Wan2.1-I2V~\cite{wan2025wan} 720p variant to support diverse generation settings, rather than the lower-resolution 480p model.
We set the LoRA~\cite{hu2022lora} rank to 128 and 256 for the Bridge and Droid configurations, respectively. To maximize data utilization, we randomly sample two views from the three available in Bridge V2. For Droid, we fix the wrist-mounted camera as the first view and select the second view from the two third-person perspectives.
% To avoid ambiguity in identity prompting under multi-view supervision, we select segmented identity references from a single view during training.

\begin{table*}[t]
\centering
\caption{
    Comparison of evaluation results on the WidowX robot in \textsc{SimplerEnv}.
    We evaluate two variants of our RoboVIP: a text-prompt–conditioned multi-view inpainting video diffusion model, and another version with additional visual identity prompting conditions (denoted as ID).
    Each task is performed on 100 trials.
    Each entry shows \textit{Grasp/Put}, where Put is the conditional success rate given a successful Grasp
    ($\text{Put} = \text{Success} / \text{Grasp}$), and the \textit{Success} column reports overall task success.
    \textbf{Bold} and \underline{underlined} numbers in the Average Success column indicate the best and second-best performance.
}
\vspace{-0.25cm}
\label{tab:widowx_results}
\setlength{\tabcolsep}{2pt}
\small
\resizebox{\textwidth}{!}{%
\begin{tabular}{c *{5}{cc}}
\toprule
\multirow{2}{*}{\textbf{Model}} &
\multicolumn{2}{c}{\shortstack{Put spoon\\on towel}} &
\multicolumn{2}{c}{\shortstack{Put carrot\\on plate}} &
\multicolumn{2}{c}{\shortstack{Stack green cube\\on yellow cube}} &
\multicolumn{2}{c}{\shortstack{Put eggplant\\in basket}} &
\multicolumn{2}{c}{Average} \\
\cmidrule(lr){2-3}\cmidrule(lr){4-5}\cmidrule(lr){6-7}\cmidrule(lr){8-9}\cmidrule(lr){10-11}
& Grasp/Put & Success & Grasp/Put & Success & Grasp/Put & Success & Grasp/Put & Success & Grasp/Put & Success \\
\midrule
% ===================== OCTO BLOCK =====================
Octo~\cite{team2024octo} (Zero-Shot) &
34\% / 26\% & 9\%  &
35\% / 20\% & 7\%  &
28\% / 0\%  & 0\%  &
65\% / 51\% & 33\% &
40.5\% / \underline{30.1\%} & 12.2\% \\

Octo (Bridge V2 SFT) &
52\% / 41\% & 29\% &
32\% / 37\% & 14\% &
47\% / 4\%  & 3\%  &
60\% / 11\% & 5\% &
\underline{47.5\%} / 23.0\% & 12.8\% \\

Octo+RoboEngine \cite{yuan2025roboengine} &
67\% / 21\% & 14\% &
43\% / 37\% & 16\% &
43\% / 5\%  & 2\%  &
0\%  / 0\%  & 0\%  &
38.2\% / 20.9\% & 8.0\%  \\

Octo+\textbf{RoboVIP} (Text prompt) &
59\% / 7\%  & 4\%  &
69\% / 46\% & 32\% &
55\% / 9\%  & 5\%  &
63\% / 17\% & 11\% &
\textbf{61.5\%} / 21.1\%& \underline{13.0\%} \\

Octo+\textbf{RoboVIP} (Text prompt with ID) &
59\% / 63\% & 37\% &
37\% / 62\% & 23\% &
47\% / 15\% & 7\%  &
37\% / 19\% & 7\%  &
45.0\% / \textbf{41.1\%} & \textbf{18.5\%} \\

\midrule % ----------- CLEAR SEPARATION ---------------
% ===================== PI_0 BLOCK =====================
$\pi_0$ \cite{black2024pi_0} (Zero-Shot) &
52\% / 63\% & 33\% &
0\%  / 0\%  & 0\%  &
28\% / 7\%  & 2\%  &
31\% / 42\% & 13\% &
27.75\% / 43.2\% & 12\%\\

$\pi_0$ (Bridge V2 SFT) &
57\% / 63\% & 36\% &
44\% / 43\% & 19\% &
30\% / 7\%  & 2\%  &
29\% / 41\% & 12\% &
40\% / 43.1\% & 17.25\% \\

$\pi_0$+RoboEngine \cite{yuan2025roboengine} &
61\% / 70\% & 43\% &
33\% / 30\% & 10\% &
61\% / 11\% & 7\%  &
31\% / 45\% & 14\% &
46.5\% / 39.8\% & 18.5\% \\

$\pi_0$+\textbf{RoboVIP} (Text prompt) &
74\% / 84\% & 62\% &
52\% / 40\% & 21\% &
49\% / 14\% & 7\%  &
36\% / 64\% & 23\% &
\underline{52.75\%} / \textbf{55.0\%} & \textbf{29\%} \\

$\pi_0$+\textbf{RoboVIP} (Text prompt with ID) &
73\% / 64\% & 47\% &
49\% / 41\% & 20\% &
52\% / 13\% & 7\%  &
53\% / 70\% & 37\% &
\textbf{56.75\%} / \underline{48.9\%} & \underline{27.75\%} \\
\bottomrule
\vspace{-0.4cm}

\end{tabular}%
}
\end{table*}

\subsection{Video Generation Results}

\input{table/generative_models_comparisons}

\begin{figure}[t]
\centering
\vspace{-0.2cm}
\includegraphics[width=1\linewidth]{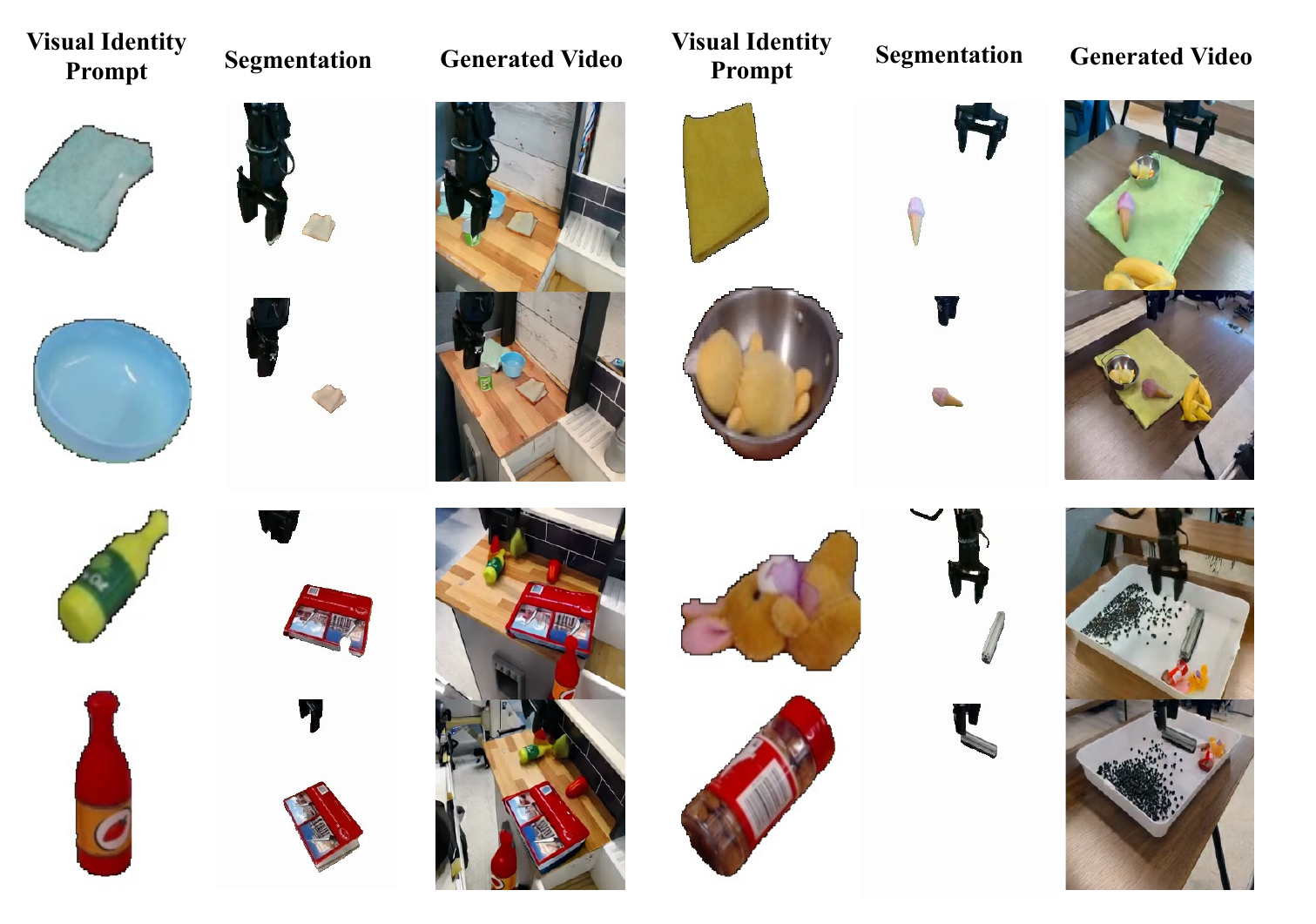}
\vspace{-0.8cm}
\caption{
\textbf{Augmented BridgeV2 Data by our RoboVIP for VLA Training}.
Our visual identity prompting enriches tabletop contents and introduces additional distractors to create more challenging settings for the policy model. The visual identity is randomly selected from our proposed pools. \textbf{Zoom in} for the best view.
}

\vspace{-0.6cm}
\label{fig:bridge_results}
\end{figure}

Our goal is to develop a scalable, plug-and-play multi-view inpainting-based generative framework that serves as an effective augmentation solution for robotic manipulation. 
To this end, we evaluate against Cosmos-Transfer2.5~\cite{ali2025world}, a video diffusion model designed for real-to-real generation, and RoboEngine~\cite{yuan2025roboengine}, an inpainting-based approach similar to ours, on the held-out test subset of the Droid~\cite{khazatsky2024droid} dataset consisting of 300 test cases.
We consider both a wrist-mounted view and a third-person view for augmentation. 
For RoboEngine, we use identical robot and object segmentation masks as our method for a fair comparison. 
For Cosmos-Transfer2.5, we evaluate its edge-conditioned variant, using its native 720p setup, and the model is conditioned on the same Qwen2.5-VL~\cite{bai2025qwen2} captioned text prompts like ours. 
For our RoboVIP, we apply visual identity prompting.
For all methods, we generate at most 49 frames per episode, starting from the first frame.

We report standard generative video metrics: Fréchet Inception Distance~\cite{heusel2017gans} (FID), Fréchet Video Distance~\cite{unterthiner2018towards} (FVD), and Learned Perceptual Image Patch Similarity~\cite{zhang2018unreasonable} (LPIPS). 
FID measures single-frame visual quality via distributional differences, while FVD captures temporal coherence and video-level dynamics. 
LPIPS quantifies image-level perceptual similarity between generated outputs and ground truth in deep feature space rather than pixel space. 
These metrics measure on vertically stitched inputs due to the multi-view setting.
To reflect the multi-view nature of our setting, we follow prior works~\cite{liu2025robotransfer, bai2024syncammaster} 
and evaluate cross-view correspondence by counting matched feature points between two generated views (MV-Mat.). 
A higher count indicates better spatial consistency and generative stability. 
We use GIM~\cite{shen2024gim} as the correspondence model, keeping all confidence thresholds and hyperparameters identical to its demo configuration.

As shown in Tab.~\ref{tab:generative_model_comparisons}, our method consistently outperforms prior approaches on most quantitative metrics. The improvement can be attributed to the fact that RoboEngine operates under a single-frame, single-view setting, while Cosmos-Transfer2.5 overlooks the requirements of multi-view generation.
In supplementary, we will also include a human study for the visual identity prompting-oriented comparisons.
%  Talk about the qualitative in this paragraph
As shown in Fig.~\ref{fig:vdm_comparison}, compared to RoboEngine, our RoboVIP performs distinguished temporal consistency. Compared to Cosmos-Transfer2.5, ours unleashes diverse scene generation, which is not limited by the pixel-aligned conditions, like edges or depth. This is thanks to our inpainting design choices. Further, none of the methods achieve multi-view consistent generation.

\subsection{Simulation Results} 
\label{sec:simulation_results}

To evaluate performance in a reproducible and scalable way, we employ the simulation environment suite SimplerEnv~\cite{li24simpler}, which shows realistic textures in the simulation environment like the real-world and has been shown to correlate well with real-world robot manipulation performance \cite{zheng2024tracevla}.
SimplerEnv enables a consistent benchmark of manipulation policies under common robot setups. 

% \noindent\textbf{VLA Implementation Details.}
We evaluate our pipelines on two recent vision-language-action models: Octo-base~\cite{team2024octo} and $\pi_0$~\cite{black2024pi_0}. Specifically, we adopt the Octo-base model as a \textbf{multi-frame conditioned} policy (with 2 history frames) and the $\pi_0$ model as a \textbf{single-frame} VLA policy. 
We fine-tune both Octo and $\pi_0$ on 8 NVIDIA GPUs with 48GB of memory each. 
All experiments use a global seed of 42 and identical data processing and augmentation settings following their official preprocessing pipeline.Both Octo and $\pi_0$ are evaluated under three training regimes:
\begin{itemize}
  \item \textbf{Zero-shot}: both Octo-base and $\pi_0$ are directly deployed without any further supervised fine-tuning in the SimplerEnv tasks.
  \item \textbf{Supervised Fine-Tuning (SFT) on BridgeDataV2}: we fine-tune each model using the open-ended instruction-conditioned dataset BridgeDataV2~\cite{walke2023bridgedata} and then deploy.
  \item \textbf{Mixed-policy baseline vs. our method}: We mix BridgeDataV2 with the augmented data produced by \textbf{RoboEngine}~\cite{yuan2025roboengine} and our proposed \textbf{RoboVIP}.
  % focusing on mixing BridgeDataV2 with additional augmented data by generative model to improve generalization performance. 
  We evaluate two variants of our multi-view inpainting video diffusion model.
  The first variant uses only text prompts as the generative condition.
  The second variant augments the same architecture with our visual identity prompting, which provides exemplar images as additional conditioning signals as shown in Fig.~\ref{fig:bridge_results}.
\end{itemize}

Tab.~\ref{tab:widowx_results} presents quantitative comparisons across the SimplerEnv tasks. 
For the Octo family, our RoboVIP (Text+ID) achieves an average success rate of 18.5\%, improving upon Octo zero-shot experiment (12.2\%) and Bridge SFT version (12.8\%), and our text-prompt-only variant (13.0\%). 
For $\pi_0$, our RoboVIP (Text-only) configuration yields the highest overall success at 29.0\%, outperforming both the SFT baseline (17.25\%) and RoboEngine (18.5\%). The Text+ID variant performs similarly at 27.75\%, confirming that both visual identity prompting and temporally consistent generative augmentation contribute to stronger generalization. 
% \vspace{-0.2\baselineskip}

A closer look at the decomposition into \emph{Grasp} and conditional \emph{Put} success (\textit{Put = Success / Grasp}) reveals the source of these gains. For Octo, our RoboVIP (Text+ID) attains the best average \emph{Put} success at 41.1\%, significantly higher than the 23.0\% achieved by Octo SFT. 
In $\pi_0$, the text-only RoboVIP obtains the highest \emph{Put} success of 55.0\%, exceeding the SFT baseline (43.1\%) and RoboEngine (39.8\%). These results indicate that our method not only improves task initiation (grasping) but also strengthens the more challenging post-grasp “Put” phase, demonstrating enhanced closed-loop control and task completion reliability.
% \vspace{-0.2\baselineskip}

The observed improvements arise from RoboVIP’s ability to generate temporally consistent, multi-view scenes that closely approximate real data distributions. For models such as Octo, which condition on multiple frames, our generated sequences provide realistic motion continuity that mitigates frame inconsistency issues present in RoboEngine. 
For $\pi_0$, the multi-view setup aligns with its pretraining configuration, reducing the gap between synthetic and real data distributions. Moreover, the use of visual identity prompts enriches scene diversity (as shown in Fig.~\ref{fig:bridge_results}) and introduces controlled clutter, which empirically benefits learning in visually complex settings. As a result, generative data from RoboVIP can closely approach—or even surpass—the effectiveness of real fine-tuning data.

\begin{figure}[t]
  \centering
  \vspace{-0.3cm}
  \includegraphics[width=0.97\linewidth]{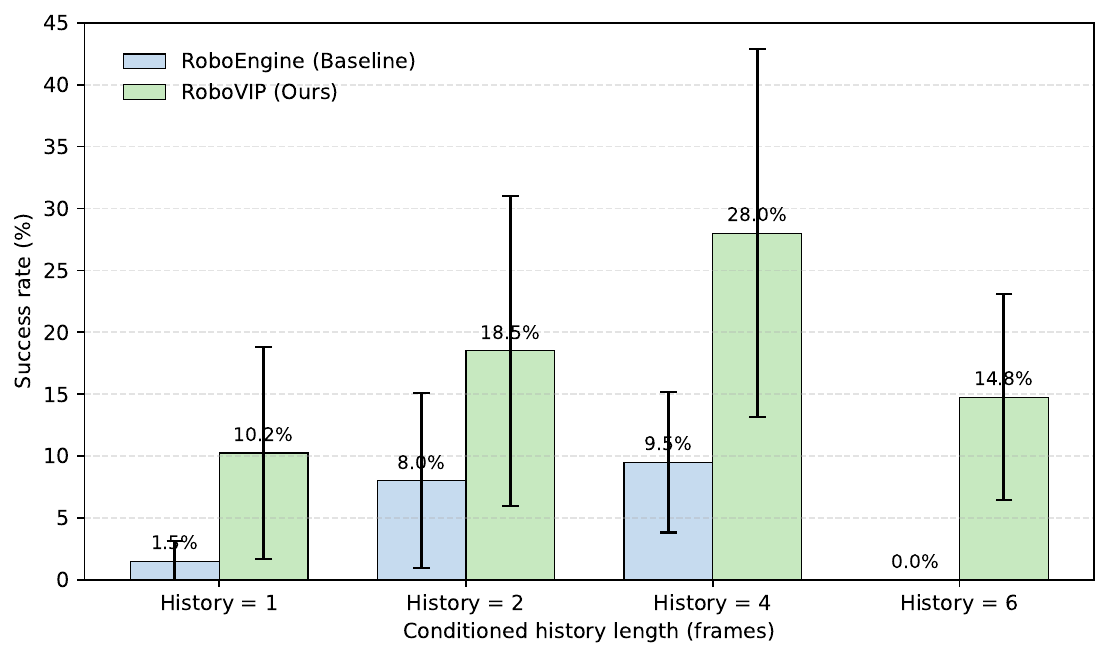}
  \vspace{-0.4cm}
  \caption{
    Policy success rate vs.\ conditioned history length (frames).
    Bars show task-averaged success rates for RoboEngine (baseline) and our RoboVIP (Text prompt with ID) on Octo~\cite{team2024octo}.
    The error bars denote standard deviation across tasks.
    The bar plot indicates that RoboVIP consistently outperforms RoboEngine, whose average success falls to zero at six history frames.
  }
  \label{fig:history-line}
  \vspace{-0.45cm}
\end{figure}

Further, we compare the influence of history length on VLA success in Fig.~\ref{fig:history-line}.
% The evaluated VLA model is Octo, which supports an arbitrary number of historical frames.
We retrained and tested Octo on different numbers of history observation frames.
Across all historical lengths, our RoboVIP maintains consistently higher success rates than RoboEngine (baseline).
Notably, while RoboEngine’s performance collapses to nearly zero under six-frame conditioning, our RoboVIP still preserves meaningful success rates, underscoring its robustness to longer temporal contexts.
This trend suggests that video-level generative augmentation—not image diffusion—is a more scalable and forward-looking direction, which is applicable for future long-horizon needs on VLA training.

\subsection{Real-World Robot Results} 
\label{sec:real_robot_results}

To specifically validate the effectiveness of our RoboVIP augmentation pipeline against real-world background distractors, we conduct experiments using a 7-DoF Franka Research 3 robotic arm equipped with a Robotiq gripper. We design a cube stacking task, which requires grasping a blue cube and stacking it onto the red cube. All experiments are conducted using Diffusion Policy (DP) \cite{chi2025diffusion}.
We established two experimental settings to test robustness against background distractors:
\textbf{Open space:} A clean background with no distractors.
\textbf{Cluttered:} A scene with 4 different distractor objects. We compare the performance of two policies:
\begin{itemize}
    \item \textbf{DP:} A DP model trained solely on 100 real-world demonstration trajectories.
    \item \textbf{DP + RoboVIP (Text+ID) :} A DP model trained on a mixed dataset of 200 trajectories, consisting of the 100 original demonstrations and 100 additional trajectories augmented by our RoboVIP framework.
\end{itemize}

\begin{figure}[t]
  \centering
  \vspace{-0.3cm}
  \includegraphics[width=\linewidth]{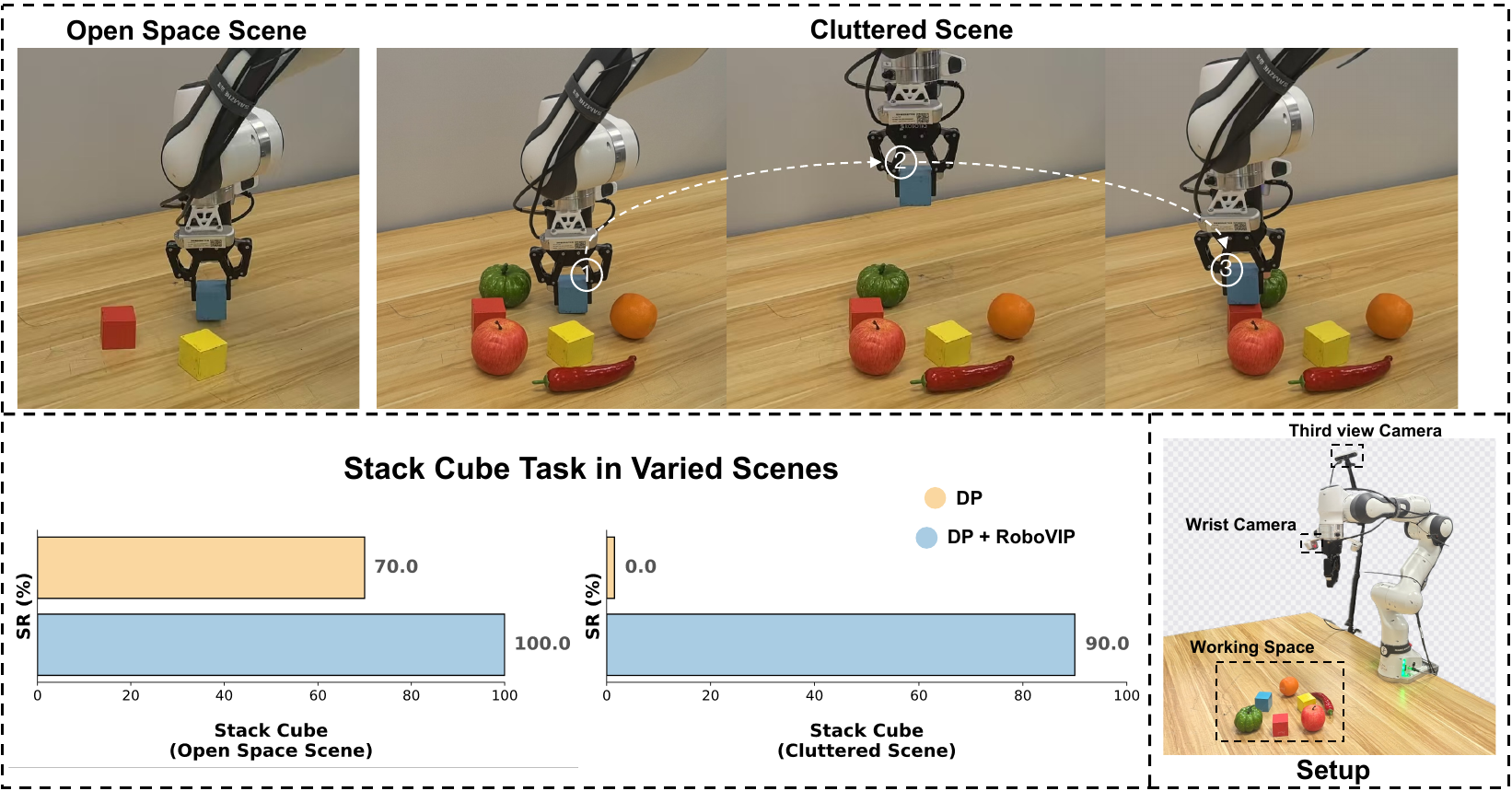}
  \vspace{-0.6cm}
  \caption{
    \textbf{Real Robot Experiment on Diffusion Policy.} Both policies were trained using identical parameter settings and measured over 10 trials.
  }
  \label{fig:real-world}
  \vspace{-0.35cm}
\end{figure}

As shown in Figure \ref{fig:real-world}, the baseline DP model's success rate drops from 7/10 in the \textbf{Open space} setting to 0/10 in the \textbf{Cluttered} setting. In contrast, the \textbf{DP + RoboVIP} model achieves perfect 10/10 success in the open space setting and maintains a robust 9/10 success rate in the cluttered setting. This demonstrates that our augmentation pipeline significantly enhances the policy's generalization and robustness to real-world visual distractors.
More details and experiments are presented in the supplementary.

%% file: table/generative_models_comparisons.tex
\begin{table}[t]
\centering
\small
\caption{
    \textbf{Generative Model Comparisons} on 300 test cases of Droid~\cite{khazatsky2024droid}. 
    Cosmos refers to Cosmos-Transfer2.5~\cite{ali2025world}.
    The best is \textbf{highlighted}.
}
\vspace{-0.2cm}
\begin{tabular}{l|c|c|c|c}
\toprule
Method & FID$\downarrow$ & FVD$\downarrow$ & LPIPS$\downarrow$ & MV-Mat.$\uparrow$ \\
\midrule
Cosmos~\cite{ali2025world} & 47.43 & 325.4 & \textbf{0.353} & 1583.4 \\
RoboEngine~\cite{yuan2025roboengine} & 62.77 & 1788.8 & 0.598 & 1301.9  \\
\textbf{RoboVIP} (Ours) & \textbf{39.97} & \textbf{138.4} & 0.409 & \textbf{2242.1} \\
\bottomrule
\end{tabular}
\vspace{-0.3cm}
\label{tab:generative_model_comparisons}
\end{table}

%% file: sec/5_conclusion.tex
\section{Conclusion}

In this work, we introduce RoboVIP, a multi-view inpainting video diffusion model with visual identity prompting to augment visual observations of the robotic manipulation data in a plug-and-play manner.
We augment large-scale data and demonstrate its effectiveness in both vision-language-action and visuomotor policy models on both the simulation environment and the real-world robot deployment.

\textbf{Limitation.}
Although our method can automate large-scale visual data augmentation and we prove its effectiveness on VLA and visuomotor policy learning, several limitations stem from current tools.
State-of-the-art video segmentation~\cite{ravi2024sam} still struggles with gripper localization and flickering; VLM reasoning~\cite{azzolini2025cosmos, bai2025qwen2} often fails to identify interactive objects; and the open-vocabulary segmentation~\cite{yuan2025roboengine, zhang2024evf} frequently produces incorrect masks and does not produce consistent results in multi-view inputs. Furthermore, although our real-world experiments leverage multi-view observations, the SimplerEnv\cite{li24simpler} benchmark only supports single-view image inputs; therefore, more extensive simulation studies are needed to fully evaluate the benefits of multi-view consistency training.

%% file: sec/X_suppl.tex
% \clearpage
% clearpage will starts from the next page

% \setcounter{page}{1}
\maketitlesupplementary

\section{Overview}
\label{sec:rationale}

This supplementary material provides more implementation and technical details, additional quantitative comparisons with analysis, and more qualitative visualization to complement the main manuscript. 
% We also recommend readers open the \textbf{HTML} website included in the ZIP package for a better check on the visualizations.
In Sec.~\ref{sec:more_experimental_detail}, we present more details of our video diffusion model, including dataset curation, caption generation, segmentation workflows, and visual identity filtering. 
We then elaborate on the simulation setup used for evaluating augmented data in tabletop manipulation tasks, followed by details of our real-world robot experiments. 
In Sec.~\ref{sec:supp_additional_results}, we provide real-world robot baseline comparisons and a user study to prove our proposed visual identity prompting. 
Finally, in Sec.~\ref{sec:supp_more_visualization}, we provide more visualizations of rollouts and generated videos to support the qualitative findings.

\section{More Experimental Details}
\label{sec:more_experimental_detail}

\subsection{More Video Generation Details}

In the dataset curation, the text prompt is composed of the scene setup description and the robot arm action description. 
The scene setup caption for VLM~\cite{bai2025qwen2} is:
\textit{Describe the scene setup in the video (exclude the robot arm) within 15 words. The input is a vertically stitched multi-view video. Only provide information with high confidence.}
The action description caption for VLM is:
\textit{Describe the action of the robot arm briefly in the video within 15 words. The input is a vertically stitched multi-view video. Only provide information with high confidence.}
Additionally, for long-duration datasets such as Droid~\cite{khazatsky2024droid}, we split the video into shorter frame chunks rather than using the entire sequence, preventing the diffusion model using text prompt from describing events that do not occur in the segment. 
The caption for VLM, instead, is:
\textit{Describe the action of the robot arm briefly in the video within 10 words. Do not predict what will happen. Just focus on what has been done. The input is a vertically stitched multi-view video. Only provide information with high confidence.}

Our segmentation pipeline is illustrated in Fig.~\ref{fig:seg_pipeline} of the main paper and consists of two parallel streams: one for robot-arm segmentation and one for interacted-object segmentation.
The gripper state annotations vary across datasets. For Bridge V1~\cite{ebert2021bridge} and V2~\cite{walke2023bridgedata}, the gripper open/close state is provided as a Boolean indicating whether the gripper is closed. For Droid~\cite{khazatsky2024droid}, the gripper open/close state is a continuous value representing the physical distance between the two fingertips. We convert this to a Boolean state through thresholding to maintain compatibility with the Bridge format.
Meanwhile, we apply a temporal buffer when converting gripper states, adding a constant offset of 5 frames before the detected gripper-close event and 5 frames after the gripper-open event. This enlarges the effective action window and mitigates annotation edge cases.
Next, open-vocab segmentation needs to input an accurate query. For the robot-arm stream, we simply use the fixed query ``robot.''
Additionally, we uniformly sample five frames from each video and select the frame with the largest robot arm masking region estimated by the segmentation model. This is because we observe that the robot arm does not always appear in the first frame or the last frame; the consideration of middle frames is needed.
For the wrist camera view, we additionally leverage off-the-shelf VLM reasoning models, Cosmos-Reason1~\cite{azzolini2025cosmos} (7B), which are trained on large-scale robotics datasets to enable plug-and-play labeling.
All videos are resized to a unified resolution of 320$\times$448 to ensure stable and consistent segmentation performance.
Afterward, we apply an outlier filter and sample segmentation keypoints using K-means, following the procedure described in the main manuscript.

% \begin{figure*}[t]
%     \centering
%     \includegraphics[width=\linewidth]{figs/figure1_supp.pdf}
%     \caption{
%     \textbf{Segmentation Pipeline.}
%     Our segmentation pipeline comprises two parallel streams: one for robot-arm segmentation and one for interacted-object segmentation. We first use the gripper-action signal to identify accurate keyframe ranges, which is helpful to locate the interacted objects that are not visible in the first or last frame (check the red bounding box inside the figure). We then leverage off-the-shelf models such as Cosmos-Reason1~\cite{azzolini2025cosmos} and SAM2~\cite{ravi2024sam}, together with several heuristic refinements, to obtain accurate masks in a fully plug-and-play manner.
%     }
%     \label{fig:seg_pipeline}
% \end{figure*}

% 这里讲ID的segmentation
In the visual identity curation, the class detected by OneFormer~\cite{jain2023oneformer} has 133 classes, but not all of them are what we need in the robotics scene. 
Thus, we manually select classes that correlate to the robotics table-top and background scene: 
\texttt{
bus, boat,
sheep, cow, elephant, bear, zebra, giraffe, backpack,
handbag, suitcase, frisbee, skis, snowboard, sports ball, baseball bat,
baseball glove, tennis racket, bottle, wine glass, cup,
spoon, bowl, banana, apple, sandwich, orange, broccoli, carrot, hot dog,
cake, chair, laptop, remote,
keyboard, cell phone, book, clock,
scissors, teddy bear, blanket, cardboard,
flower, fruit, pillow, towel, food-other-merged,
microwave, oven, toaster, refrigerator, potted plant, couch,
banner, net, platform, bench,
mirror-stuff, bed, vase, tent,
paper-merged, cabinet-merged, curtain%
}.
We crop each visual identity into a square bounding box and mask the background. We discard identities whose masked background covers more than 60\% of the bounding box, as overly large masks often lead to visually abnormal or distorted identity crops.

% Quality Filtering估计最后整理code的时候要全部调整，很多filtering的criteria估计会用的不同后面
To curate the visual identity pool, we apply a series of quality-control filters.  
First, we assess overall image quality using CLIP-IQA~\cite{wang2023exploring} and remove the lowest 50\%.  
Second, to exclude extremely small or disproportionately large objects, we discard crops falling in the smallest and largest 10\% by resolution size.  
Third, we measure image sharpness using the variance of the Laplacian on grayscale images and filter out the lowest 30\%.  
Finally, for CLIP~\cite{radford2021learning} text–image alignment, we remove 60\% of samples with the weakest similarity scores.

In the video diffusion model training, the learning rate is set to 5e-5 with 100 steps of warm-up. 
We use 8-bit Adam to save GPU memory in the training.
In the training, the maximum number of frames is set to 33 for Droid~\cite{khazatsky2024droid} and 49 for Bridge~\cite{walke2023bridgedata}, which presents similar training time. This is because most Bridge datasets are shorter than 49 frames, but almost all Droid datasets are longer than 49 frames.
Since our pretrained model, Wan2.1~\cite{wan2025wan}, has a CLIP-based image encoder for the first frame, we directly set that with our first segmented frame without doing other modifications.
The frames are sampled in their original fps without applying acceleration like general video generation~\cite {yang2024cogvideox, wan2025wan}.
Our visual-identity input is randomly resized within a scale range of $[0.8, 1.2]$ to avoid fixed scale learning for the visual identity reference. To reduce computation during both training and inference, we use at most one packed identity frame. 
An additional frame consumes computation equivalent to four temporal frames in Wan2.1~\cite{wan2025wan} due to its temporal compression ratio of 4, and the attention cost scales quadratically with sequence length ($O(N^2)$). 
If an instance does not require visual identity prompting, we simply omit this identity frame rather than inserting any padding as placeholder.

For the evaluation of RoboEngine on the Droid dataset, we apply the same segmentation masks condition as in our method. 
RoboEngine does not provide an automatic caption system and assumes ideal text-prompt inputs by the user; however, using VLM-generated captions may introduce hallucinated descriptions of background elements such as tables. 
For example, the prompt might be ``put the apple on the table''. Their system will directly input the word ``table'' into the open-vocab segmentation model.
If large background regions (e.g., the table) are mistakenly included in the segmented video as conditions, the diffusion model would be exposed to an unrealistically large portion of ground-truth pixels, thereby inflating the bias in the evaluation.
Compared to largely modifying their segmentation workflow, we choose to directly use our segmented results as their condition, which induces a more direct comparison of the generative model capability. 
For the Bridge~\cite{walke2023bridgedata} augmentation used in simulation, we directly use the task descriptions provided by the dataset on RoboEngine. These captions are short, concise, and generally more reliable than those for Droid, enabling a fair comparison of end-to-end augmentation capabilities.

\begin{figure*}[t]
    \centering
    \includegraphics[width=\linewidth]{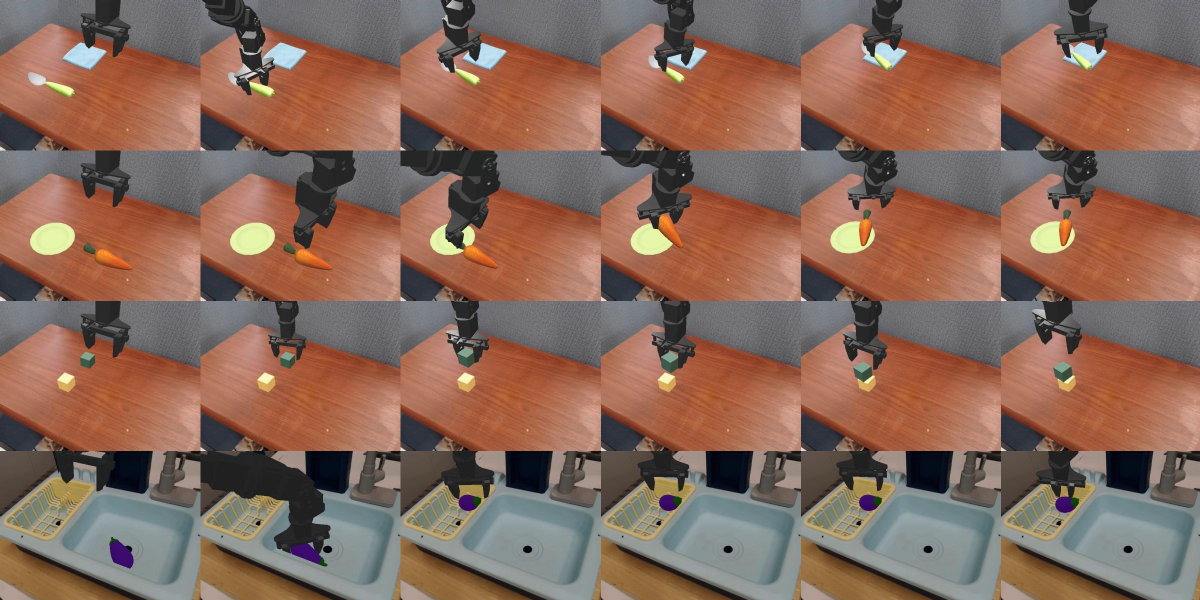}
    \caption{\textbf{$\pi_0$ rollouts in SimplerEnv.}
    Visualization of rollouts for the $\pi_0$ policy on the same four tasks
    and in the same order as Figure~\ref{fig:octo_rollout}. 
    As before, each row corresponds to a single task and shows frames sampled
    uniformly in time from left (initial state) to right (final state). 
    Comparing this figure with Figure~\ref{fig:octo_rollout} provides a qualitative
    view of the differences in behavior and convergence between the two policies.}
    \label{fig:pi0_rollout}
\end{figure*}
\begin{figure*}[t]
    \centering
    \includegraphics[width=\linewidth]{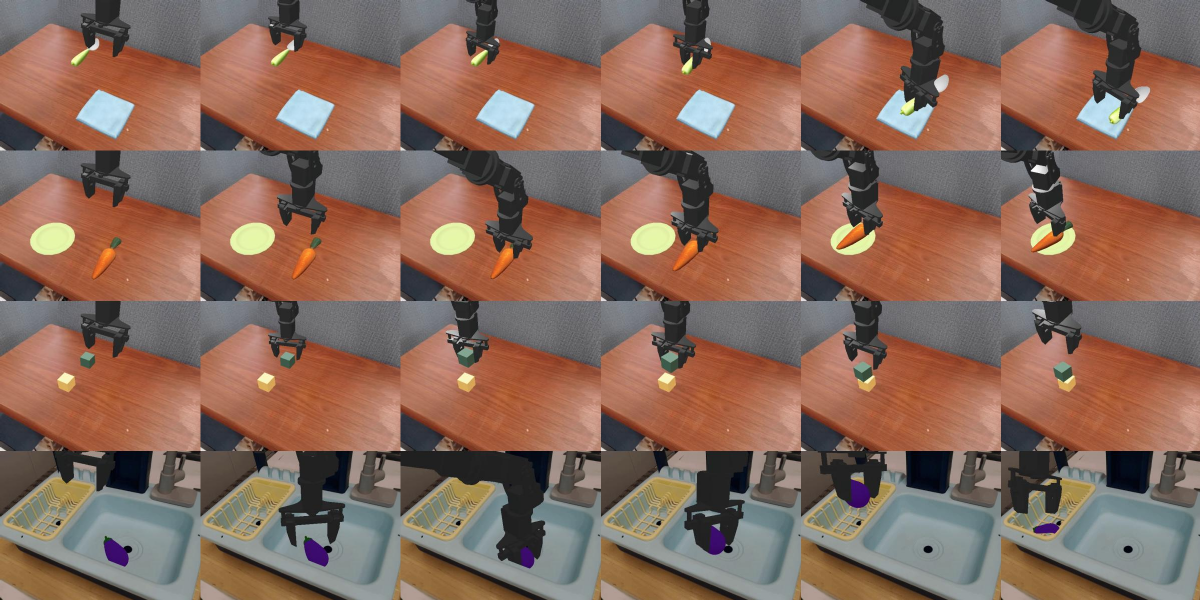}
    \caption{\textbf{Octo rollouts in SimplerEnv.}
    Each row shows a rollout of the Octo policy on one task. 
    From top to bottom: \texttt{spoon\_on\_towel}, \texttt{carrot\_on\_plate}, 
    \texttt{stack\_cube}, and \texttt{put\_eggplant\_in\_basket}. 
    Within each row, frames are sampled uniformly in time from an episode
    and arranged from left (start of the rollout) to right (end of the rollout),
    illustrating how the policy gradually moves the object toward the goal region.}
    \label{fig:octo_rollout}
\end{figure*}

\subsection{More Simulation Setup Details}

As discussed in Sec.~\ref{sec:simulation_results}, we conduct experiments on fine-tuning vision--language--action (VLA) models. To demonstrate the effectiveness of our data augmentation method, we evaluate the performance of the fine-tuned models on the SimplerEnv simulation benchmark~\cite{li24simpler}. 

SimplerEnv provides lightweight tabletop manipulation environments that closely mirror the scenes in BridgeData V2~\cite{walke2023bridgedata}. Each environment consists of a fixed camera observing a planar workspace, a WidowX 250 6-DoF robot arm, and a small number of colored objects or receptacles (e.g., towel, plate, cubes, basket). The robot starts from a reset joint configuration, with the target object placed in a random but task-consistent location on the table, and the target receptacle placed in a distinct region of the workspace. 
Overall success is defined by the final spatial relationship between the manipulated object and its target receptacle (e.g., on top of, inside). Grasp success is defined by the target object is grasped during execution.

Since our data augmentation method is applied to the BridgeData V2 dataset, which is collected using a WidowX 250 6-DoF robot arm, we conduct our evaluation in the corresponding SimplerEnv environments that emulate these real-world scenes. The four tasks we consider are summarized below.

\textbf{1. Put Spoon on Tablecloth}:
The scene contains a metallic spoon and a blue cloth patch (towel) placed on the tabletop. At the start of each episode, the spoon is initialized at a random position away from the cloth. The goal is to grasp the spoon and place it such that it lies on top of the blue cloth region, indicating successful completion of the task.

\textbf{2. Put Carrot on Plate}:
This environment includes an orange carrot object and a green plate on the table. The carrot is initially positioned off the plate, with random variation in its pose and distance to the target. The objective is to pick up the carrot and place it so that it rests on the surface of the green plate, without requiring a specific orientation.

\textbf{3. Stack Green Block on Yellow Block}:
The scene consists of two cubic blocks: a green block and a yellow block. The yellow block serves as the base and is placed in a fixed region of the workspace, while the green block starts at a separate, randomly sampled position. The goal is to precisely place the green block on top of the yellow block, forming a stable stack. This task requires both accurate vertical placement and alignment of the cube centers.

\textbf{4. Put Eggplant in Basket}:
In this setting, the tabletop contains a purple eggplant-shaped object and an open basket. The eggplant is initialized outside the basket at varying positions, while the basket remains in a fixed area. The robot must grasp the eggplant and place it inside the basket volume (rather than merely touching the rim). Note that BridgeData V2 does not contain an exact instruction and demonstration set for this specific ``eggplant in basket'' configuration; instead, SimplerEnv provides a scene that is only semantically related to the original data. As a result, this task primarily evaluates the policy's ability to generalize to a novel but conceptually similar goal.

Since augmenting a large-scale multi-view video dataset is computationally expensive, we subsample episodes using task-relevant keywords for the main manuscript comparisons. Specifically, we filter episodes using the keywords \texttt{Spoon}, \texttt{Cloth}, \texttt{Carrot}, \texttt{Cube}, \texttt{Eggplant}, and \texttt{Basket}, resulting in a curated set of 12k episodes. 
For Octo SFT, we evaluate the 80k checkpoint, and for fine-tuning with mixed data (e.g., with RoboEngine and RoboVIP) we use the 100k checkpoint to ensure the same number of epochs. For $\pi_0$, we use the 20k checkpoint for supervised fine-tuning and the 25k checkpoint for evaluation. Since the amount of data is increased by roughly 20\%, we also increase the number of training iterations by approximately 20\% to enable a fair comparison. For evaluation in SimplerEnv, we only evaluate each policy on a single static camera view, since SimplerEnv does not provide a wrist-camera view.

During Octo training, we enable its built-in data augmentation. Specifically, we use a uniform goal relabeling strategy with a subsample length of 100, and a task augmentation strategy that deletes task conditioning while keeping each image with probability 0.5. 
All Octo-Base models are trained on eight Nvidia GPUs with 48\,GB memory each and a global batch size of 128. 
The same hardware setup is used for $\pi_0$, and all other training hyperparameters follow the defaults in their official repositories.

We also experimented with LIBERO~\cite{liu2023libero} and ManiSkill3~\cite{taomaniskill3}. 
However, we found that the off-the-shelf segmentation model~\cite{ravi2024sam, zhang2024evf} performed poorly in these simulators. Furthermore, these simulation environments have texture distributions that differ significantly from those in our real-world training data.
% The stylized and less photorealistic scenes in LIBERO and ManiSkill3 lead to inaccurate object masks, which in turn make our data augmentation procedure unreliable and uninformative. 
In contrast, SimplerEnv is designed to closely match the real-world scenes in BridgeData V2, and thus provides visual distributions that are better aligned with policies trained on real robot data. For this reason, we focus our quantitative evaluation on SimplerEnv.

\begin{table}[t]
\centering
\small
\caption{
    \textbf{Generative Model Comparisons on Real-World Experiments.} 
    Success rates averaged over 10 trials.
    The best is \textbf{highlighted}.
    % DP refers to Diffusion Policy~\cite{chi2025diffusion}.
} 
%\vspace{-0.2cm}
\begin{tabular}{l|c|c}
\toprule
Method & Open Space & Cluttered  \\
\midrule
Diffusion Policy~\cite{chi2025diffusion} (DP) & 7/10& 0/10  \\
DP + RoboEngine~\cite{yuan2025roboengine} & 8/10 & 2/10  \\
DP + Cosmos-Transfer2.5~\cite{ali2025world} & 3/10 & 3/10  \\
\textbf{RoboVIP} (Ours) & \textbf{10/10} & \textbf{9/10} \\
\bottomrule
\end{tabular}
%\vspace{-0.3cm}
\label{tab:real_world_results}
\end{table}

\subsection{Real-World Robot Setup}
% Put environment setup like other papers

\noindent\textbf{Experiment Environments.} We conduct all real-world experiments using a 7-DoF Franka Research 3 robotic arm equipped with a Robotiq gripper. The table also features a multi-camera system, comprising a wrist-mounted Intel RealSense D435 and a third-person Intel RealSense D455. Both cameras capture RGB frames with a resolution of $640 \times 480$ at 30 fps.

\noindent\textbf{Training Data.} Demonstrations were gathered via teleoperation using a 3Dconnexion SpaceMouse. Frankcontroller is utilized for low-level communication with the robot and gripper during both data collection and policy evaluation. Our dataset for the open-space experiment comprises 100 real human demonstrations, each episode has nearly 180 frames. In addition, we synthesized 100 augmented demonstrations based on these curated trajectories using our proposed RoboVIP and each baseline method. For the comparison of augmentation methods, we constructed training sets of 200 episodes by combining the synthesized demonstrations with the original real-world data.

\noindent\textbf{Data Augmentation.} 
For our real-world robot data augmentation by video diffusion model, video conditions are split into chunks of up to 33 frames before synthesis. Chunks shorter than 33 frames are padded to the nearest $4N+1$ length required by the temporal compression scheme of the video diffusion model.

\noindent\textbf{Policy Training.} We train and evaluate RoboVIP and 2 additional demonstration augmentation methods, including RoboEngine~\cite{yuan2025roboengine} and Cosmos-Transfer2.5~\cite{ali2025world}, on enhancing the vanilla Diffusion Policy~\cite{chi2025diffusion}. For the experiment setting without augmentation methods, the models are trained for 2500 steps, and 5000 steps for all augmented settings. The batch size is set to 64. We utilize Lerobot~\cite{cadene2024lerobot} as the dataloader and trainer. During the data preprocessing, we resize all the different views of observations to 224$\times${224} and downsample the videos to 10 FPS during both the training process and evaluation as the model input. We also filter a few episodes that have a total frame length than 300. During training, we set the horizon, observation steps, and action steps of Diffusion Policy~\cite{chi2025diffusion} to 8, 2, and 4 for all the models. When training with RoboEngine and Cosmos-Transfer2.5, we keep the default settings reported in their manuscripts.

\noindent\textbf{Policy Evaluation.}
We designed a \textit{Cube Stacking} task to evaluate the policy's precision and robustness. The objective is to grasp a target blue cube and stack it onto a red cube. A trial is considered successful only if the blue cube is stably placed atop the red cube. A trial is recorded as a failure if any of the following occur: (1) the robot fails to grasp the blue cube; (2) the cube is dropped during transport; or (3) the blue cube is not successfully placed on the red cube or fails to remain stable after the gripper releases it. 

We propose two different settings for our real-world experiments, including \textbf{Open space}: A clean background with no distractors. and \textbf{Cluttered}: A scene with 4 different unseen distractor objects.

\section{Additional Results and Analysis}
\label{sec:supp_additional_results}

\subsection{Real-World Robot Baseline Comparisons}

% We recommend readers check our website to see more dynamic and realistic videos. 
% \textbf{Augmented video.} It is worth noting that all generative model augmented methods exhibit some degree of flickering in the robot's motion during real-world rollouts. This appears to be a broader limitation of current generative model-driven augmentation pipelines, suggesting that achieving smoother and more temporally stable action generation on visuomotor policy learning remains an open challenge for future work.
\textbf{Real-World Experiment Results.}
We extend the experiments presented in Sec.~\ref{sec:real_robot_results} to compare our RoboVIP against two additional generative methods. The quantitative results, measured by success rates over 10 trials, are listed in Tab.~\ref{tab:real_world_results}. 
The results demonstrate that our RoboVIP achieves the best performance in enhancing Diffusion Policy across both seen tasks and unseen environments. Notably, Diffusion Policy trained with RoboVIP attains a significantly higher success rate in the Cluttered setting (\textbf{9/10}) compared to other augmentation methods that rarely succeed, while maintaining stable performance under normal and unseen settings.

\begin{figure*}[t]
    \centering
    \vspace{0.3cm}
    \includegraphics[width=\linewidth]{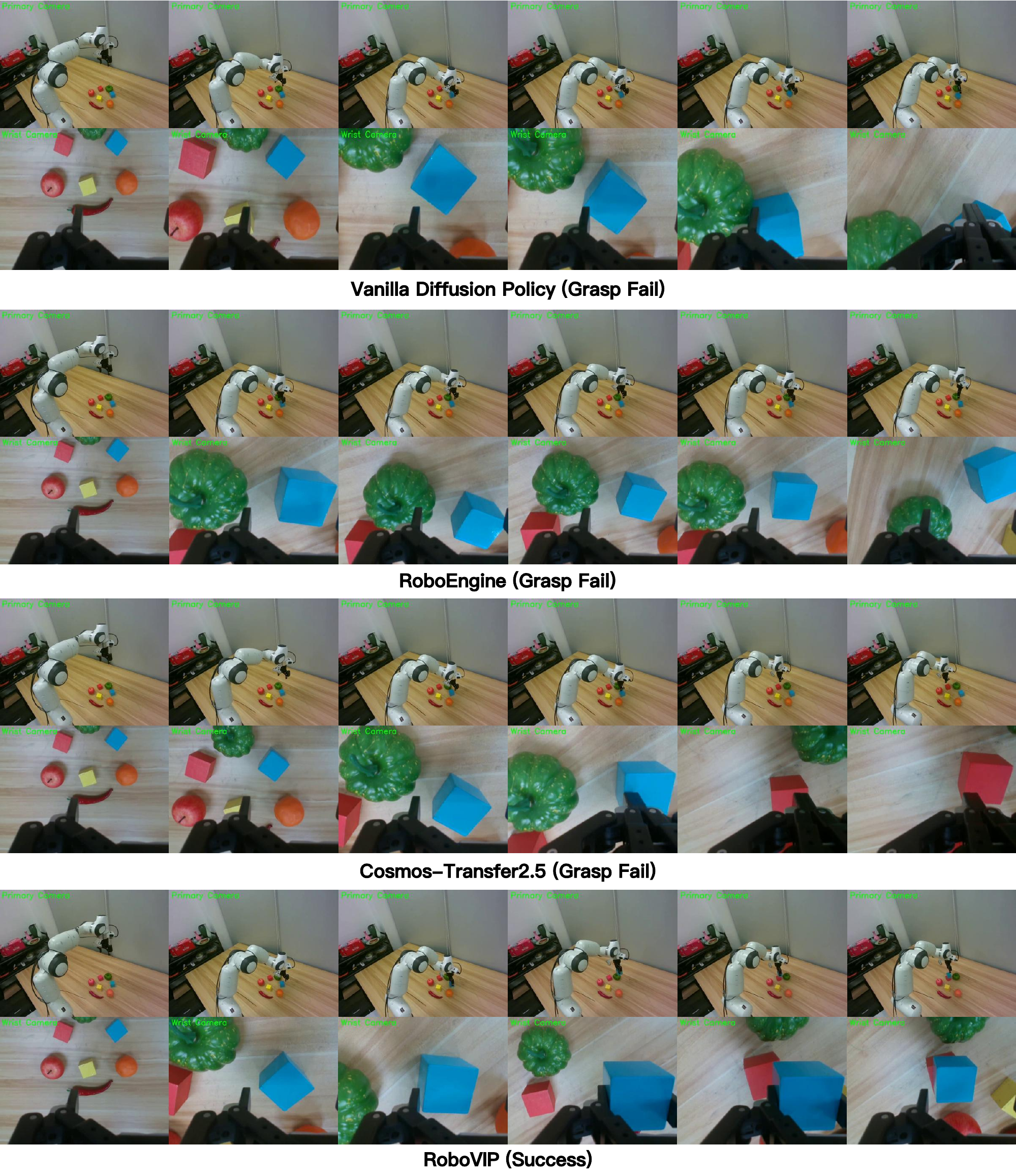}
    \caption{
    \textbf{Real-World Robot Rollout Results.}
    Baseline methods~\cite{chi2025diffusion, yuan2025roboengine, ali2025world} fail during the grasping stage under the cluttered setup, whereas our approach achieves a successful grasp and final placement.
    More samples are available on our website.
    }
    \label{fig:real_rollout}
\end{figure*}

\noindent\textbf{Visualizations.}
We visualize the policy rollouts in the cluttered scene setting. 
As shown in Fig.~\ref{fig:real_rollout}, baseline methods—including vanilla Diffusion Policy and policies trained with Cosmos-Transfer2.5~\cite{ali2025world} or RoboEngine~\cite{yuan2025roboengine} augmentations—struggle to generalize in the presence of background distractors.
Specifically, vanilla Diffusion Policy and {RoboEngine} frequently fail to correctly localize the target object, resulting in grasp failures on the blue cube, while Cosmos-Transfer2.5 fails to place the cube stably. In contrast, the policy augmented by our {RoboVIP} demonstrates robust generalization capabilities. It successfully executes the full horizon of the task: accurately grasping the blue cube amidst distractors, stably transporting it, and precisely stacking it onto the red cube without the cube falling.

\subsection{User Study on Visual Identity Prompting}

Since our visual identities correspond to small tabletop objects, it is difficult to rely on DINO~\cite{simeoni2025dinov3} or CLIP-based~\cite{radford2021learning} metrics, as in general video generation~\cite{liu2025phantom, fei2025skyreels} to reliably retrieve these small objects' features amid the overwhelming background semantics.
To more faithfully demonstrate how our proposed visual identity prompting enriches the generated tabletop scenes, we conduct an additional user study comparing our visual-identity-conditioned model against our own text-only variant.
We randomly select 50 videos for each rater and asked three anonymous human raters
to perform pairwise comparisons between two methods.
The instruction provided to the raters is:

\textit{You will be shown two videos generated by two different models, along with a visual identity image. For each comparison, please answer the following two questions and select the option that best matches the intended criterion:
1. Which video more faithfully incorporates the provided visual identity image into the scene?
2. Which video presents a tabletop that is richer in visual content?}

For Question~(1), raters preferred the visual-identity–conditioned generation in \textbf{97.3\%} of the comparisons, indicating a strong advantage in identity preservation.  
For Question~(2), visual identity prompting was preferred in \textbf{80.0\%} of the comparisons, showing that this feature encourages richer and more detailed tabletop content in the augmentation.  
These results demonstrate that conditioning on visual identity prompting not only improves identity alignment but also leads to more complex and content-rich scene compositions.

\section{More Visualization}
\label{sec:supp_more_visualization}

\subsection{Rollout Visualization}

To complement the quantitative success rates reported in Sec.~\ref{sec:simulation_results} of the main manuscript, we visualize representative rollouts of the fine-tuned policies in SimplerEnv~\cite{li24simpler}. 
In Fig.\ref{fig:pi0_rollout} and  Fig.\ref{fig:octo_rollout}, each row corresponds to one of the four evaluation tasks, and each image is a temporal strip constructed by sampling frames uniformly from the beginning to the end of an episode. Thus, the rollout progresses from left to right within each row, showing how the robot moves the object from its initial pose toward the target configuration.

\subsection{Real-World Robot Augmented Videos}

\begin{figure*}[t]
    \centering
    \includegraphics[width=\linewidth]{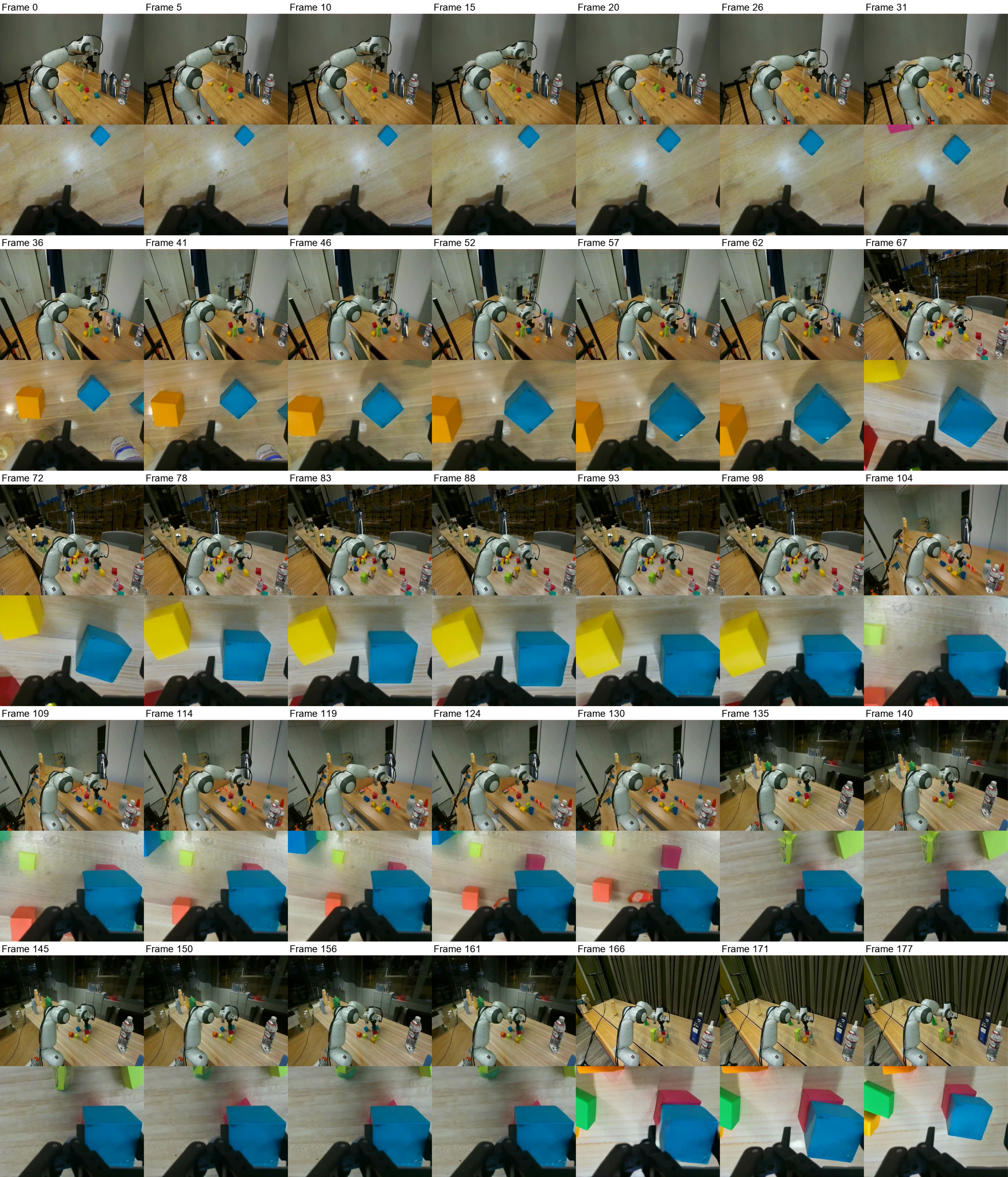}
    \caption{
    \textbf{Real-World Robot Zero-Shot Long-Horizon Augmentation by our RoboVIP.}
    We showcase long-horizon real-world video augmented by our proposed RoboVIP. 
    Since real robot videos contain far more frames than what current video diffusion models can process directly, we split the video into 33-frame chunks and generate them sequentially with different visual identities selected. 
    The figure presents equally sampled frames across the full trajectory.
    More samples are available on our website.
    }
    \label{fig:supp_real_aug}
\end{figure*}

As shown in Fig.~\ref{fig:supp_real_aug}, we include long-horizon generation results by our RoboVIP on real-world robot episodes collected in our lab.
These generated videos are directly used for downstream visuomotor policy training, Diffusion Policy~\cite{chi2025diffusion}. It is worth noting that the hardware configuration in our lab cannot be perfectly aligned with the Droid~\cite{khazatsky2024droid} training data. In particular, while Droid provides gripper views that are almost perfectly centered and symmetric, our wrist-mounted camera exhibits a noticeable left-shift bias. As a result, these real-world rollouts constitute genuine zero-shot cases for our video diffusion model.

Beyond this domain mismatch, our model demonstrates richer tabletop and scene variations under zero-shot deployment. For example, if you closely inspect the wrist-view generations, you will notice that the tabletop textures continuously change across videos, and these textures appear highly realistic. Furthermore, in the third-person view, each generation presents a different background configuration, and even the geometry of the table varies from chunk to chunk.
Thanks to our proposed visual identity prompting, the generated tabletop scenes become significantly more enriched with diverse and realistic objects.
We recommend readers check our website for more vivid videos.

\clearpage